%% file: main.tex
\documentclass{article}

\usepackage[final, nonatbib]{nips_2016}

\usepackage[utf8]{inputenc} 
\usepackage[T1]{fontenc}    
\usepackage{hyperref}       
\usepackage{url}            
\usepackage{booktabs}       
\usepackage{amsfonts}       
\usepackage{nicefrac}       
\usepackage{microtype}      
\usepackage{bm}
\usepackage{color}
\usepackage{csquotes}
\usepackage{graphicx}
\usepackage{amssymb}
\usepackage{amsmath}
\usepackage{latexsym}
\usepackage{algpseudocode}

\newcommand{\quotes}[1]{``#1''}
\usepackage[tight]{subfigure}
\usepackage{placeins}
\usepackage{algorithm}
\usepackage{algorithmicx}

\usepackage{subfigure}
\usepackage{bm}
\usepackage{xspace}

\usepackage{paralist}
\usepackage{siunitx}

\usepackage{tikz}
\usepackage{pgfplots}
\pgfplotsset{compat=1.9}
\usepackage{verbatim}
\usepackage{textcomp}
\usetikzlibrary{shapes,arrows,positioning}
\usetikzlibrary{patterns,calc}
\pgfdeclarelayer{background}
\pgfdeclarelayer{foreground} 
\pgfsetlayers{background,main,foreground}
\usepackage[normalem]{ulem}

\input{./notations}
\usepackage{lineno,hyperref}
\title{Deep Neural Networks Regularization for Structured Output Prediction}

\author{
  Soufiane Belharbi\thanks{https://sbelharbi.github.io} \\
  Normandie Univ, UNIROUEN, UNIHAVRE,\\
  INSA Rouen, LITIS\\
  76000 Rouen, France \\
  \texttt{soufiane.belharbi@insa-rouen.fr} \\
  \And
  Romain Hérault\\
  Normandie Univ, UNIROUEN, UNIHAVRE,\\
  INSA Rouen, LITIS\\
  76000 Rouen, France \\
  \texttt{romain.herault@insa-rouen.fr} \\
  \And
  Clément Chatelain \\
  Normandie Univ, UNIROUEN, UNIHAVRE,\\
  INSA Rouen, LITIS\\
  76000 Rouen, France \\
  \texttt{clement.chatelain@insa-rouen.fr} \\
  \And
  Sébastien Adam\\
  Normandie Univ, UNIROUEN, UNIHAVRE,\\
  INSA Rouen, LITIS\\
  76000 Rouen, France \\
  \texttt{sebastien.adam@univ-rouen.fr} \\
}

\begin{document}

\maketitle

\input{0-abstract}

\input{1-intro}

\input{2-relatedw}

\input{3-proposedmethod}

\input{4-implement}

\input{5-exps}

\input{6-conclusion}

\FloatBarrier
\subsubsection*{Acknowledgments}

This work has been partly supported by the grant ANR-11-JS02-010 LeMon, the grant ANR-16-CE23-0006 \quotes{Deep in France} and has benefited from computational means from CRIANN, the contributions of which are greatly appreciated.


\bibliography{bibliography}  





\end{document}

%% file: notations.tex
\newcommand{\x}{\mathbf{x} \xspace}
\newcommand{\y}{\mathbf{y} \xspace}
\newcommand{\w}{\mathbf{w} \xspace}

\newcommand{\xTilde}{\mathbf{\tilde{x}} \xspace}
\newcommand{\yTilde}{\mathbf{\tilde{y}} \xspace}
\newcommand{\xHat}{\mathbf{\hat{x}} \xspace}
\newcommand{\yHat}{\mathbf{\hat{y}} \xspace}
\newcommand{\mapSup}{\mathbf{\mathcal{M}} \xspace}
\newcommand{\mapIn}{\mathbf{\mathcal{R}_{in}} \xspace}
\newcommand{\mapOut}{\mathbf{\mathcal{R}_{out}} \xspace}

\newcommand{\sS}{{\mathcal{S}}}

\newcommand{\sL}{{\mathcal{L}}}
\newcommand{\sF}{{\mathcal{F}}}

\newcommand{\xest}{\hat{\x}}
\newcommand{\yest}{\hat{\y}}
\newcommand{\xproj}{\tilde{\x}}
\newcommand{\yproj}{\tilde{\y}}

\newcommand{\m}{\mathcal{M} \xspace}
\newcommand{\J}{\mathcal{J} \xspace}
\newcommand{\data}{\mathcal{D} \xspace}
\newcommand{\R}{\mathcal{R} \xspace}
\newcommand{\cost}{\mathcal{C} \xspace}

\DeclareMathOperator{\card}{card}

%% file: 0-abstract.tex
\begin{abstract}
  A deep neural network model is a powerful framework for learning representations. Usually, it is used to learn the relation $\x \to \y$ by exploiting the regularities in the input $\x$. In structured output prediction problems, $\y$ is multi-dimensional and structural relations often exist between the dimensions. The motivation of this work is to learn the output dependencies that may lie in the output data in order to improve the prediction accuracy.
  Unfortunately, feedforward networks are unable to exploit the relations between the outputs. In order to overcome this issue, we propose in this paper a regularization scheme for training neural networks for these particular tasks using a multi-task framework. Our scheme aims at incorporating the learning of the output representation $\y$ in the training process in an unsupervised fashion while learning the supervised mapping function $\x \to \y$. 
  \\
  We evaluate our framework on a facial landmark detection problem which is a typical structured output task. We show over two public challenging datasets (LFPW and HELEN) that our regularization scheme improves the generalization of deep neural networks and accelerates their training. The use of unlabeled data and label-only data is also explored, showing an additional improvement of the results. We provide an opensource  implementation\footnote{\url{https://github.com/sbelharbi/structured-output-ae}} of our framework.
\end{abstract}

%% file: 1-intro.tex
\section{Introduction}
\label{intro}
\begin{figure}[!htbp]
     \centering
     \subfigure{
     \includegraphics[width=0.2\linewidth]{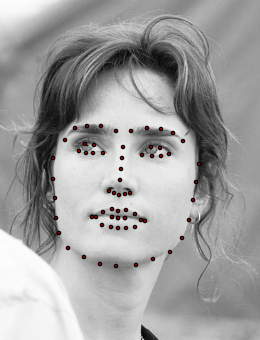}
     }
     \subfigure{
     \includegraphics[width=0.2\linewidth]{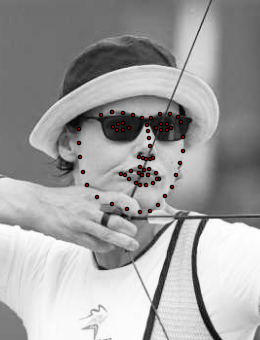}
     }
     \subfigure{
     \includegraphics[width=0.2\linewidth]{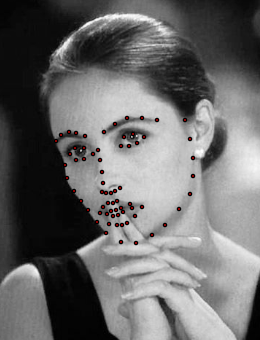}
     }
     \subfigure{
     \includegraphics[width=0.2\linewidth]{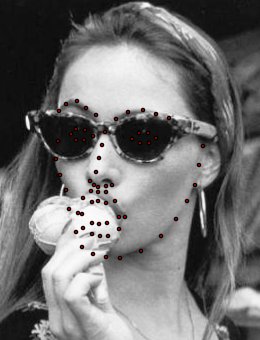}
     }
     \caption{Examples of facial landmarks from LFPW \cite{belhumeur11} 
   training set.}
     \label{fig:fig0}
\end{figure}

In machine learning field, the main task usually consists in learning general regularities over the input space in order to provide a specific output. Most of machine learning applications aim at predicting a single value: a label for classification or a scalar value for regression. Many recent applications address challenging problems where the output lies in a multi-dimensional space describing discrete or continuous variables that are most of the time  interdependent. 
A typical example is speech recognition, where the output label is a sequence of characters which are interdependent, following the statistics of the considered language.
These dependencies generally constitute a regular structure such as a sequence, a string, a tree or a graph. As it provides constraints that may help the prediction, this structure should be either discovered if unknown, or integrated in the learning algorithm using prior assumptions. The range of applications that deal with structured output data is large. One can cite, among others, image labeling \cite{farabet13, longSD15, nohHH15, ronnebergerFB15, ZhangDG17, HoffmanWYD16, LiUBBSB16, SohnLY15}, statistical natural language processing (NLP) \cite{JaderbergSVZ14b2014, och03, sleator95, schmid94}, bioinformatics \cite{jones99, syed09}, speech processing \cite{rabiner89, zen09} and handwriting recognition \cite{graves2009novel,Stuner16}. Another example which is considered in the evaluation of our proposal in this paper is the facial landmark detection problem. The task consists in predicting the coordinates of a set of keypoints given the face image as input (Fig.\ref{fig:fig0}). The set of points are interdependent throughout geometric relations induced by the face structure. Therefore, facial landmark detection can be considered as a structured output prediction task.

One main difficulty in structured output prediction is the exponential number of possible configurations of the output space. From a statistical point of view, learning to predict accurately high dimensional vectors requires a large amount of data where in practice we usually have limited data. In this article we propose to consider structured output prediction as a representation learning problem, where the model must i) capture the discriminative relation between $\x$ (input) and $\y$ (output), and ii) capture the interdependencies laying between the variables of each space by efficiently modeling the input and output  distributions. We address this modelization through a regularization scheme for training neural networks. Feedforward neural networks lack exploiting the structural information between the $\y$ components. Therefore, we incorporate in our framework an unsupervised task which aims at discovering this hidden structure. The advantage of doing so is there is no need to fix beforehand any prior structural information. The unsupervised task learns it on itself.

Our contributions is a multi-task framework dedicated to train feedforward neural networks models for structured output prediction. We propose to combine unsupervised tasks over the input and output data in parallel with the supervised task. This parallelism can be seen as a regularization of the supervised task which helps it to generalize better. Moreover, as a second contribution, we demonstrate experimentally the benefit of using the output labels $\y$ without their corresponding inputs $\x$. In this work, the multi task framework is instantiated using auto-encoders \cite{vincent10, bengio07} for both representations learning and exploiting  unlabeled data (input) and label-only data (output). We demonstrate the efficiency of our proposal over a real-world facial landmark detection problem.

The rest of the paper is organized as follows. Related works about structured output prediction is proposed in section \ref{relatedw}. Section \ref{sec:mtl} presents the proposed formulation and its optimization details. Section \ref{sec:impl} describes the instantiation of the formulation using a deep neural network. Finally, section \ref{sec:expes} details the conducted experiments including the datasets, the evaluation metrics and the general training setup. Two types of experiments are explored: with and without the use of unlabeled data. Results are presented and discussed for both cases.

%% file: 2-relatedw.tex
\section{Related work}
\label{relatedw}
We distinguish two main categories of methods for structured output prediction.  For a long time, graphical models have showed a large success in different applications involving 1D and 2D signals. Recently, a new trend has emerged based on deep neural networks.

\subsection{Graphical Models Approaches}
Historically, graphical models are well known to be suitable for learning structures. One of their main strength is an easy integration of explicit structural constraints and prior knowledge directly into the model's structure. They have shown a large success in modeling structured data thanks to their capacity to capture dependencies among relevant random variables. For instance, Hidden Markov Models (HMM) framework has a large success in modeling sequence data. HMMs make an assumption that the output random variables are supposed to be independent which is not the case in many real-world applications where strong relations are present. Conditional Random Fields (CRF) have been proposed to overcome this issue, thanks to its capability to learn large dependencies of the observed output data. These two frameworks are widely used to model structured output data represented as a 1-D sequence \cite{elyacoubi02,rabiner89,bikel99,lafferty01}. Many approaches have also been proposed to deal with 2-D structured output data as an extension of HMM and CRF. \cite{nicolas06} propose a Markov Random Field (MRF) for document image segmentation. \cite{szummer04} provide an adaptation of CRF to 2-D signals with hand drawn diagrams interpretation. Another extension of CRF to 3-D signal is presented in \cite{tsechpenakis07} for 3-D medical image segmentation. Despite the large success of graphical models in many domains, they still encounter some difficulties. For instance, due to their inference computational cost, graphical models are limited to low dimensional structured output problems. Furthermore, HMM and CRF models are generally used with discrete output data where few works address the regression problem \cite{noto12,fridman93}.

\subsection{Deep Neural Networks Approaches}

More recently, deep learning based approaches have been widely used to solve structured output prediction, especially proposed for image labeling problems. Deep learning domain provides many different architectures. Therefore, different solutions were proposed depending on the application in hand and what is expected as a result.

In image labeling task (also known as semantic segmentation), one needs models able to adapt to the large variations in the input image. Given their large success in image processing related tasks \cite{krizhevsky12}, convolutional neural networks is a natural choice. Therefore, they have been used as the core model in image labeling problems in order to learn the relevant features. They have been used either combined with simple post-processing in order to calibrate the output \cite{ciresanGGS12} or with more sophisticated models in structure modeling such as CRF \cite{farabet13} or energy based models \cite{ningDLPBB05}. Recently, a new trend has emerged, based on the application of convolution  \cite{longSD15, ronnebergerFB15} or deconvolutional \cite{nohHH15} layers in the output of the network which goes by the name of fully convolutional networks and showed successful results in image labeling. Despite this success, these models does not take in consideration the output representation.

In many applications, it is not enough to provide the output prediction, but also its probability. In this case, Conditional Restricted Boltzmann Machines, a particular case of neural networks and probabilistic graphical models have been used with different training algorithms according to the size of the plausible output configurations  \cite{mnihLH11}. Training and inferring using such models remains a difficult task. In this same direction, \cite{belangerM16} tackle structured output problems as an energy minimization through two feed-forward networks. The first is used for feature extraction over the input. The second is used for estimating an energy by taking as input the extracted features and the current state of the output labels. This allows learning the interdependencies within the output labels. The prediction is performed using an iterative backpropagation-based method with respect to the labels  through the second network which remains computationally expensive. Similarly, Recurrent Neural Networks (RNN) are a particular architecture of neural networks. They have shown a great success in modeling sequence data and outputing sequence probability for applications such as Natural Language Processing (NLP) tasks \cite{liu2014, sutskeverVL14, auliGQZ13} and speech recognition \cite{gravesJ14}. It has also been used for image captioning \cite{karpathyL15}. However, RNN models doe not consider explicitly the output dependencies.

In \cite{lerouge15}, our team proposed the use of auto-encoders in order to learn the output distribution in a pre-training fashion with application to image labeling with promising success. The approach consists in two sequential steps. First, an input and output pre-training is performed in an unsupervised way using autoencoders. Then, a finetune is applied on the whole network using supervised data. While this approach allows incorporating prior knowledge about the output distribution, it has two main issues. First, the alteration of a network output layer is critical and must be performed carefully. Moreover, one needs to perform multiple trial-error loops in order to set the autoencoder's training hyper-parameters. The second issue is overfitting. When pre-training the output auto-encoder, there is actually no information that indicates if the pre-training is helping the supervised task, nor when to stop the pre-training.

The present work proposes a general and easy to use multi-task training framework for structured output prediction models. The input and the output unsupervised tasks are embedded into a regularization scheme and learned in parallel with the supervised task. The rationale behind is that the unsupervised tasks should provide a \textit{generalization} aspect to the main supervised task and should limit overfitting. This parallel transfer learning which includes an output reconstruction task constitutes the main contribution of this work. In structured output context, the role of the output task is to learn the hidden structure within the original output data, in an unsupervised way. This can be very helpful in models that do not consider the relations between the components of the output representation such as feedforward neural networks. We also show that the proposed framework enables to use labels without input in an unsupervised fashion and its effect on the generalization of the model. This can be very useful in applications where the output data is abundant such as in a speech recognition task where the output is ascii text which can be easily gathered from Internet. In this article, we validate our proposal on a facial landmark prediction problem over two challenging public datasets (LFPW and HELEN). The performed experiments show an improvement of the generalization of deep neural networks and an acceleration of their training.

%% file: 3-proposedmethod.tex
\section{Multi-task Training Framework for Structured Output Prediction}
\label{sec:mtl}
Let us consider a training set $\data$ containing examples with both features and targets $(x,y)$,  features without target~$(x,\_)$, and targets without features~$(\_,y)$.
Let us consider a set $\sF$ which is the subset of $\data$ containing examples with at least features~$x$, a set $\sL$ which is the subset of  $\data$ containing examples with at least targets~$y$, and a set $\sS$ which is the subset of  $\data$ containing examples with both features~$x$ and targets~$y$.
One can note that all examples in $\sS$ are also in $\sF$ and in $\sL$ .

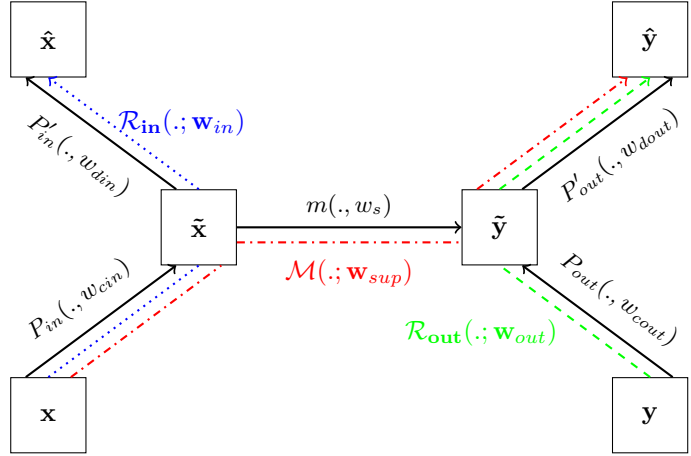
\begin{figure}[h!]
	\begin{center}
		\input{framework}
	\end{center}
	\caption{Proposed MTL framework. Black plain arrows stand for intermediate functions, blue dotted arrow for input auxiliary task $\R_{in}$, green dashed arrow for output auxiliary task $\R_{out}$, and red dash-dotted arrow for the main supervised task $\m$.}
	\label{fig:fig0.1}
\end{figure}

\begin{description}
	
\item[Input task] \hfill \\
The input task $\R_{in}$ is an unsupervised reconstruction task which  aims at learning global and more robust input representation based on the original input data $\x$.
This task projects the input data $\x$ into an intermediate representation space $\xproj$ through a coding function $P_{in}$, known as encoder. Then, it attempts to recover the original input by reconstructing $\xest$ from $\xproj$ through a decoding function $P'_{in}$, known as decoder:
\begin{equation}
\xest=\R_{in}\left(\x; \w_{in}\right) =P'_{in}\left(\xproj=P_{in}\left(\x;\w_{cin}\right);\w_{din}\right) \; ,
\end{equation}

\noindent where $\w_{in}=\{\w_{cin}, \w_{din}\}$.
The decoder parameters $\w_{din}$ are proper to this task however the encoder parameters $\w_{cin}$ are shared with the main task (see Fig.\ref{fig:fig0.1}). This multi-task aspect will attract, hopefully, the shared parameters in the parameters space toward regions that build more general and robust input representations and avoid getting stuck in local minima. Therefore, it promotes generalization. This can be useful to start the training process of the main task.

The training criterion for this task is given by :

\begin{equation}
  \label{eq:eq2}
  \J_{in}(\sF; \w_{in}) = \frac{1}{\card{\sF}}\sum\limits_{x \in
    \sF}\cost_{in}(\R_{in}(\x; \w_{in}) , \x) \; ,
\end{equation}
\noindent where $\cost_{in}$ is an unsupervised learning cost which can be computed on all the samples with features (i.e. on $\sF$).  Practically, it can be the mean squared error.

\item[Output task] \hfill \\
The output task $\R_{out}$ is an unsupervised reconstruction task which has the same goal as the input task.
Similarly, this task projects the output data $\y$ into an intermediate representation space $\yproj$ through a coding function $P_{out}$, i.e. a coder. Then, it attempts to recover the original output data by reoncstructing  $\yest$ based on $\yproj$ through a decoding function $P'_{out}$, i.e. a decoder. In structured output data, $\yproj$ can be seen as a code that contains many aspect of the original output data $\y$, most importantly, its hidden structure that describes the global relation between the components of $\y$. This hidden structure is discovered in an unsupervised way without  priors fixed beforehand which makes it simple to use. Moreover, it allows using labels only (without input $\x$) which can be helpful in tasks with abundant output data such as in speech recognition task (Sec.\ref{relatedw}):
\begin{equation}
\yest=\R_{out}\left(\y; \w_{out}\right) =P'_{out}\left(\yproj=P_{out}\left(\y;\w_{cout}\right);\w_{dout}\right) \enspace.
\end{equation}

\noindent where $\w_{out} = \{\w_{cout}, \w_{dout}\}$.
In the opposite of the input task, the encoder parameters $\w_{cout}$ are proper to this task while the decoder parameters $\w_{dout}$ are shared with the main task (see Fig.\ref{fig:fig0.1}).

The training criterion for this task is given by :
\begin{equation}
  \label{eq:eq3}
  \J_{out}(\sL; \w_{out}) = \frac{1}{\card{\sL}}\sum\limits_{y \in \sL}\cost_{out}(\R_{out}(\y;
  \w_{out}) , \y) \; ,
\end{equation}
\noindent where $\cost_{out}$ is an unsupervised learning cost which can be computed on all the samples with labels (i.e. on $\sL$), typically, the mean squared error.

\item[Main task] \hfill \\
The main task is a supervised task that attempts to learn the mapping function $\m$ between features $\x$ and labels $\y$.
In order to do so, the first part of the mapping function is shared with the encoding part $P_{in}$ of the input task and the last part is shared with the decoding part $P'_{out}$ of the output task. The middle part $m$ of the mapping function $\m$ is specific to this task:
\begin{equation}
\yest=\m\left(\x; \w_{sup}\right) =P'_{out}\left(m\left(P_{in}\left(\x;\w_{cin}\right);\w_{s}\right);\w_{dout}\right) \enspace.
\end{equation}
\noindent where  $\w_{sup}=\{\w_{cin}, \w_{s}, \w_{dout}\}$. Accordingly, $\w_{cin}$ and $\w_{dout}$ parameters are respectively shared with the input and output tasks.

Learning this task consists in minimizing its learning criterion $\J_s$, 
\begin{equation}
  \label{eq:eq1}
  \J_{s}(\sS; \w_{sup}) = \frac{1}{\card{\sS}}
  \sum\limits_{(x, y) \in \sS}\cost_{s}(\m(x; \w_{sup}) , y) \; ,
\end{equation}
\noindent where $\cost_{s}(\cdot, \cdot)$ can be the mean squared error.
\end{description}

\bigskip

\FloatBarrier

As a synthesis, our proposal is formulated as a multi-task learning framework (MTL) \cite{caruana97ML}, which gathers a main task and two secondary tasks. This framework is illustrated in Fig. \ref{fig:fig0.1}.

Learning the three tasks is performed in parallel.
This can be translated in terms of training cost as the sum of the corresponding costs.
Given that the tasks have different importance, we weight each cost using a corresponding importance weight $\lambda_{sup}$, $\lambda_{in}$ and $\lambda_{out}$ respectively for the supervised, the input and output tasks.
Therefore, the full objective of our framework can be written as:
\begin{equation}
  \label{eq:eq4}
  \J(\data; \w)  =  \lambda_{sup} \cdot \J_{s}(\sS; \w_{sup})
  + \lambda_{in} \cdot \J_{in}(\sF; \w_{in})
  +\lambda_{out} \cdot \J_{out}(\sL; \w_{out})\; ,
\end{equation}
\noindent where $\w = \{ \w_{cin}, \w_{din}, \w_{s}, \w_{cout}, \w_{dout}\}$ is the complete set of parameters of the framework.

Instead of using fixed importance weights that can be difficult to optimaly set, we evolve them through the learning epochs. In this context, Eq. \ref{eq:eq4} is modified as follows :
\begin{align}
\label{eq:eq5}
\J(\data; \w)  &=  \lambda_{sup}(t) \cdot \J_{s}(\sS; \w_{sup}) + \lambda_{in}(t) \cdot \J_{in}(\sF; \w_{in})
+\lambda_{out}(t) \cdot \J_{out}(\sL; \w_{out})\; ,
\end{align}
\noindent where $t \ge 0$ indicates the learning epochs. Our motivation to evolve the importance weights is that we want to use the secondary tasks to start the training and avoid the main task to get stuck in local minima early in the beginning of the training by moving the parameters towards regions that generalize better. Then, toward the end of the training, we drop the secondary tasks by annealing their importance toward zero because they are no longer necessary for the main task. The early stopping of the secondary tasks is important in this context of mult-tasking as shown in \cite{zhang14ECCV} otherwise, they will overfit, therefore, they will harm the main task. The main advantage of Eq.\ref{eq:eq5} is that it allows an interaction between the main supervised task and the secondary tasks. Our hope is that this interaction will promote the generalization aspect of the main task and prevent it from overfitting.

%% file: framework.tex
\begin{tikzpicture}

\tikzstyle{state} = [rectangle,draw=black,minimum width=10mm,minimum height=10mm];

\coordinate (csw) at (0,0);
\coordinate (cne) at (80mm,50mm);
\coordinate (cse) at (csw-|cne);
\coordinate (cnw) at  (cne-|csw);
\path (csw)--(cse) coordinate[pos=0.25] (axp) coordinate[pos=0.75] (ayp);
\path (cse)--(cne) coordinate[pos=0.5] (omid);

\node[state] (x) at (csw) {$\x$};
\node[state] (xproj) at (omid-|axp) {$\xTilde$};
\node[state] (xest) at (cnw) {$\xHat$};

\node[state] (y) at (cse) {$\y$};
\node[state] (yproj) at (omid-|ayp) {$\yTilde$};
\node[state] (yest) at (cne) {$\yHat$};

\path[postaction={transform canvas={xshift=-3mm},draw,->,thick}] (x.north)--(xproj.south) node[pos=0.5,sloped,above,yshift=3mm,xshift=-3mm] {\footnotesize $P_{in}(.,w_{cin})$};
\path[draw,thick,dotted,blue] (x.north)--(xproj.south);
\path[postaction={transform canvas={xshift=3mm},draw,thick,dashdotted,red}] (x.north)--(xproj.south);

\path[postaction={transform canvas={xshift=-3mm},draw,->,thick}] (xproj.north)--(xest.south) node[pos=0.5,sloped,below,yshift=-3mm,xshift=-3mm] {\footnotesize $P'_{in}(.,w_{din})$};
\path[draw,thick,dotted,blue,->] (xproj.north)--(xest.south) node[pos=0.6,right,blue] {$\mapIn(.; \w_{in})$};

\path[postaction={transform canvas={xshift=3mm},draw,->,thick}] (y.north)--(yproj.south) node[pos=0.5,sloped,above,yshift=3mm,xshift=3mm] {\footnotesize $P_{out}(.,w_{cout})$};
\path[draw,thick,dashed,green] (y.north)--(yproj.south) node[pos=0.4,left=3mm,green] {$\mapOut(.; \w_{out})$};

\path[postaction={transform canvas={xshift=3mm},draw,->,thick}] (yproj.north)--(yest.south) node[pos=0.5,sloped,below,yshift=-3mm,xshift=3mm] {\footnotesize $P'_{out}(.,w_{dout})$};
\path[draw,thick,dashed,green,->] (yproj.north)--(yest.south) ;
\path[postaction={transform canvas={xshift=-3mm},draw,thick,dashdotted,red,->}] (yproj.north)--(yest.south);

\draw[->,thick] (xproj)--(yproj) node[pos=0.5,above] {\footnotesize $m(.,w_{s})$};
\path[postaction={transform canvas={yshift=-2mm},draw,thick,dashdotted,red}] (xproj)--(yproj) node[pos=0.5,below=3mm,red] {$\mapSup(.; \w_{sup})$};


\end{tikzpicture}

%% file: 4-implement.tex
\section{Implementation}
\label{sec:impl}

In this work, we implement our framework throughout a deep neural network.
The main supervised task is performed using a deep neural network (DNN) with $K$ layers.
Secondary reconstruction tasks are carried out by auto-encoders (AE): the input task is achieved using an AE that has $K_{in}$ layers in its encoding part, with an encoded representation of the same dimension as $\xproj$. Similarly, the output task is achieved using an AE that has $K_{out}$ layers in its decoding part, with an encoded representation of the same dimension as $\yproj$.
At least one layer must be dedicated in the DNN to link $\xproj$ and $\yproj$ in the intermediate spaces. Therefore, $K_{in} + K_{out} < K$.

Parameters $\w_{in}$ are the parameters of the whole input AE, $\w_{out}$ are the parameters of the whole output AE and $\w_{sup}$ are the parameters of the main neural network (NN).
The encoding layers of the input AE are tied to the first layers of the main NN, and the decoding layers of the output AE are in turn tied to the last layers of the main NN.
If $\w_i$ are the parameters of layer $i$ of a neural network, then $\w_1$ to $\w_{K_{in}}$ parameters of the input AE are shared with $\w_1$ to $\w_{K_{in}}$ parameters of the main NN.
Moreover, if $\w_{-i}$ are the parameters of last minus $i-1$ layer of a neural network, then parameters $\w_{-K_{out}}$ to $\w_{-1}$ of the output AE are shared with the parameters  $\w_{-K_{out}}$ to $\w_{-1}$ of the main NN.

During training, the loss function of the input AE is used as $ \J_{in}$, the loss function of the output AE is used as $ \J_{out}$, and the loss function of the main NN is used as $\J_{s}$.

Optimizing Eq.\ref{eq:eq5} can be performed using Stochastic Gradient Descent. In the
case of task combination, one way to perform the optimization is to alternate between the tasks when needed \cite{collobert08ICML, zhang14ECCV}. In the case where the training set does not contain unlabeled data, the optimization of Eq.\ref{eq:eq5} can be done in parallel over all the tasks. When using unlabeled data, the gradient for the whole cost can not be computed at once. Therefore, we need to split the gradient for each sub-cost according to the nature of the samples at each mini-batch.
For the sake of clarity, we illustrate our optimization scheme in Algorithm \ref{alg:alg0} using on-line training (i.e. training one sample at a time). Mini-batch training can be performed in the same way.

\begin{algorithm}
    \caption{Our training strategy for one epoch}
    \label{alg:alg0}
    \resizebox{1.\textwidth}{!}{%
\begin{minipage}{1.2\textwidth}
    \begin{algorithmic}[1]
      \State $\data$ is the \emph{shuffled}
      training set. $B$ a sample.
      \For{$B$ in $\mathcal{D}$}
        \If{$B$ contains $\x$}
        	\State \texttt{Update $\w_{in}$}: Make a gradient step toward $\lambda_{in} \times \J_{in}$ using $B$ (Eq.\ref{eq:eq2}).
        \EndIf
        \If{$B$ contains $\y$}
        \State \texttt{Update $\w_{out}$}: Make a gradient step toward $\lambda_{out} \times \J_{out}$ using $B$ (Eq.\ref{eq:eq3}).
        \EndIf
        \State \texttt{\# parallel parameters update}
        \If{$B$ contains $\x$ and $\y$}
        \State \texttt{Update $\w$}: Make a gradient step toward $\J$ using $B$
        (Eq.\ref{eq:eq5}).
        \EndIf
	\State \texttt{Update $\lambda_{sup}$, $\lambda_{in}$ and $\lambda_{out}$}.
      \EndFor
    \end{algorithmic}
    \end{minipage}
}
\end{algorithm}

%% file: 5-exps.tex
\section{Experiments} 
\label{sec:expes}
 We evaluate our framework on a facial landmark detection problem which is typically a structured output problem since the facial landmarks are spatially  inter-dependent. 
 Facial landmarks are a set of key points on human face images as shown in Fig.  \ref{fig:fig0}. Each key point is defined by the  coordinates $(x,y)$ in the image ($x, y \in \mathbb{R}$). The number of landmarks is dataset or application dependent.

It must be emphasized here that the purpose of our experiments in this paper was not to outperform the state of the art in facial landmark detection but to show that learning the output dependencies helps improving the performance of DNN on that task. Thus, we will compare a model with/without input and output training. \cite{zhang14} use a cascade of neural networks. In their work, they provide the performance of their first global network. Therefore, we will use it as a reference to compare our performance (both networks has close architectures) except they use larger training dataset.

We first describe the datasets followed by a description of the evaluation metrics used in facial landmark problems. Then, we present the general setup of our experiments followed by two types of experiments: without and with unlabeled data. An opensource implementation of our MTL deep instantiation is available online\footnote{\url{https://github.com/sbelharbi/structured-output-ae}}.

\subsection{Datasets}
 
\mbox{} We have carried out our evaluation over two challenging public datasets
 for facial landmark detection problem: LFPW \cite{belhumeur11} and HELEN
 \cite{le12}.
 
 \textbf{LFPW dataset} consists of 1132 training images and 300 test images
 taken under unconstrained conditions (in the wild) with large variations in the
 pose, expression, illumination and with partial occlusions
 (Fig.\ref{fig:fig0}). This makes the facial point detection a challenging task
 on this dataset. From the initial dataset described in LFPW
 \cite{belhumeur11}, we use only the 811 training images and the 224 test
 images provided by the ibug website\footnote{300 faces in-the-wild challenge
   \url{http://ibug.doc.ic.ac.uk/resources/300-W/}}. Ground truth annotations of
 68 facial points are provided by \cite{sagonas13}. We divide the available
 training samples into two sets: validation set (135 samples) and training set
 (676 samples).
 
\begin{figure}[!htbp]
   \centering
   \subfigure{
   \includegraphics[height=0.2\linewidth]{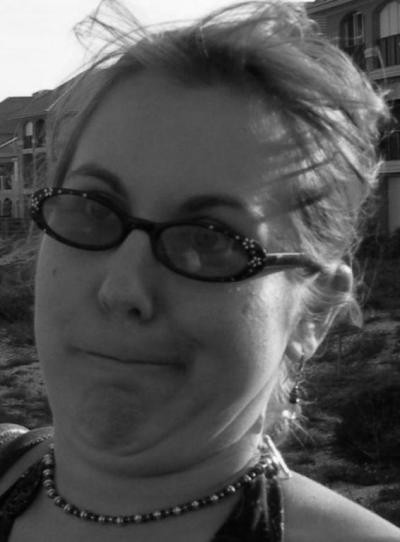}
   }
   \subfigure{
   \includegraphics[height=0.2\linewidth]{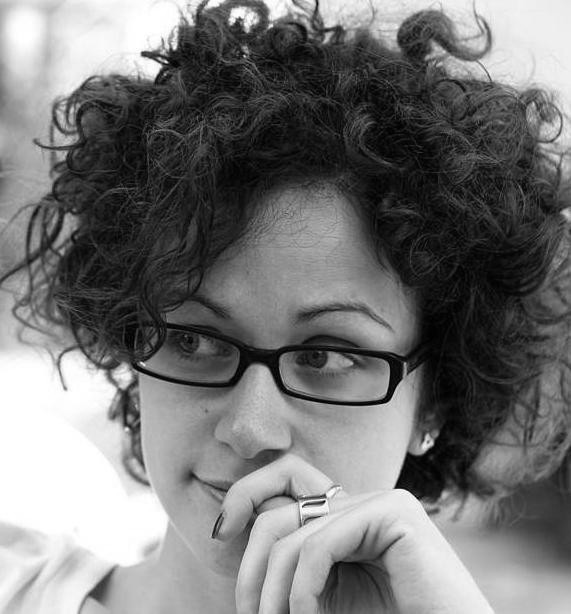}
   }
   \subfigure{
   \includegraphics[height=0.2\linewidth]{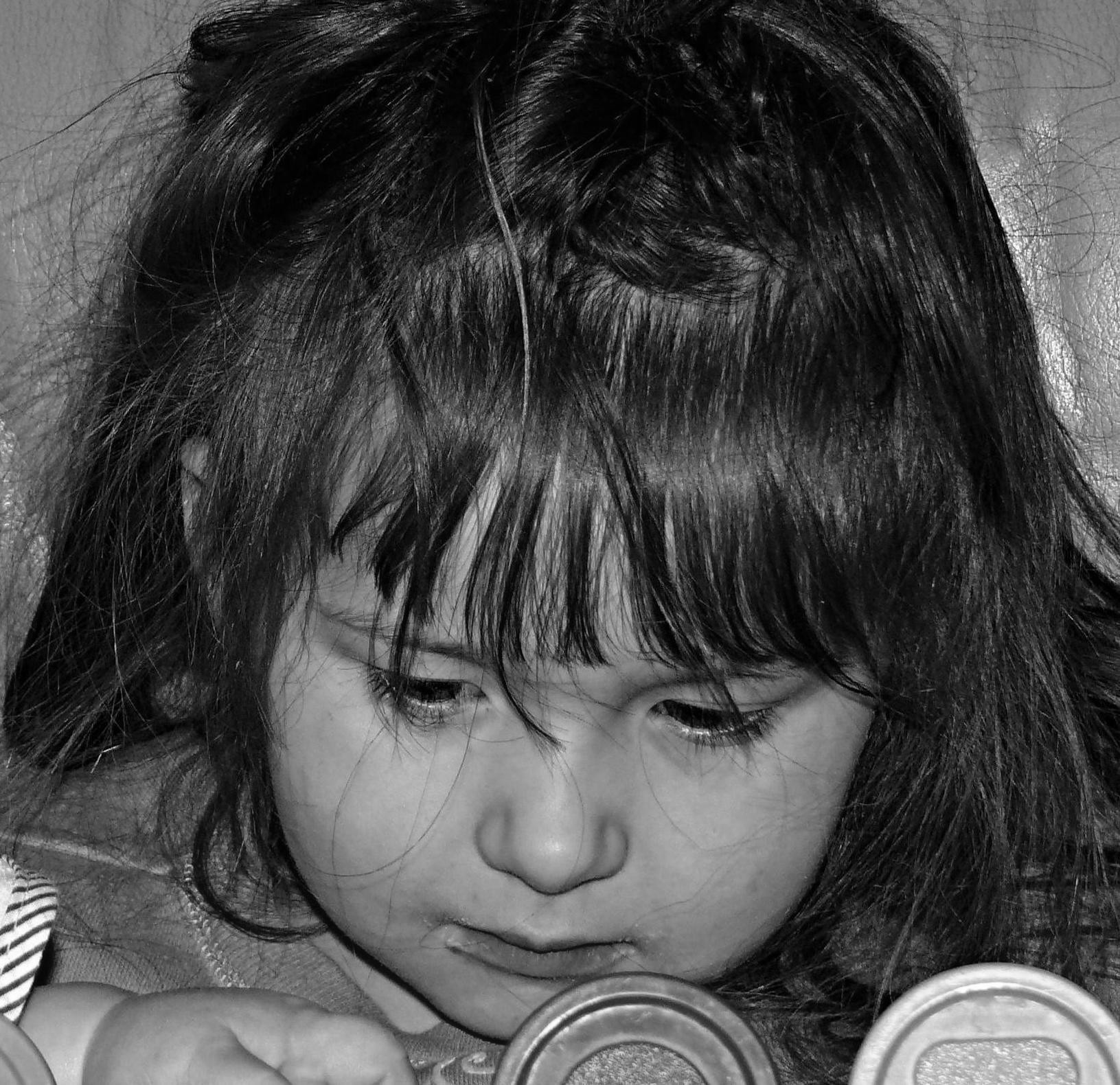}
   }
   \subfigure{
   \includegraphics[height=0.2\linewidth]{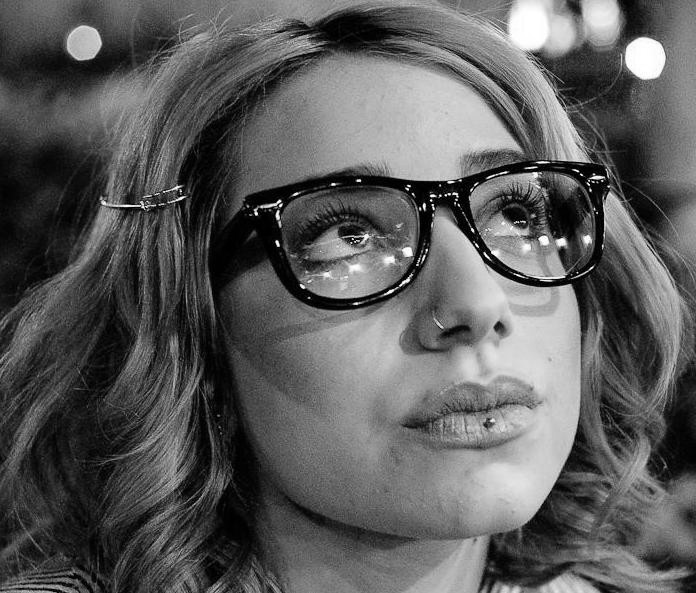}
   }
   \caption{Samples from HELEN \cite{le12} dataset.}
   \label{fig:fig1}
   \end{figure}

 \textbf{HELEN dataset} is similar to LFPW dataset, where the images have
 been taken under unconstrained conditions with high resolution and collected
 from Flikr using text queries. It contains 2000 images for training, and 330
 images for test. Images and face bounding boxes are provided by the
 same site as for LFPW. The ground truth annotations are provided by
 \cite{sagonas13}.  Examples of dataset are shown in Fig.\ref{fig:fig1}.
 
 All faces are cropped into the same size ($50 \times 50$) and pixels are normalized in [0,1]. The facial landmarks are normalized into [-1,1].

  \subsection{Metrics}
 In order to evaluate the prediction of the model, we use the standard metrics
 used in facial landmark detection problems.
 
 The Normalized Root Mean Squared Error (NRMSE)\cite{cristinacce06}
 (Eq.\ref{eq:eq22}) is the Euclidean distance between the predicted shape and the
 ground truth normalized by the product of the number of points in the shape and
 the inter-ocular distance $D$ (distance between the eyes pupils of the ground
 truth),
 \begin{equation}\label{eq:eq22}
  NRMSE(s_p, s_g) = \frac{1}{N*D} \sum^N_{i=1} ||s_{pi} - s_{gi}||_2 \enspace ,
 \end{equation}
 where $s_p$ and $s_g$ are the predicted and the ground truth shapes,
 respectively. Both shapes have the same number of points $N$.  $D$ is the
 inter-ocular distance of the shape $s_g$.
 
 Using the NMRSE, we can calculate the Cumulative Distribution Function for a
 specific NRMSE ($CDF_{NRMSE}$) value (Eq.\ref{eq:eq33}) overall the database,
 \begin{equation}\label{eq:eq33}
  CDF_x = \frac{CARD(NRMSE \le x)}{n} \enspace ,
 \end{equation}
 where $CARD(.)$ is the cardinal of a set. $n$ is the total number of images.
 
 The $CDF_{NRMSE}$ represents the percentage of images with error less or equal
 than the specified NRMSE value. For example a $CDF_{0.1}=0.4$ over a test set
 means that $40\%$ of the test set images have an error less or equal than
 $0.1$. A CDF curve can be plotted according to these $CDF_{NRMSE}$ values by
 varying the value of $NRMSE$.
 
 These are the usual evaluation criteria used in facial landmark detection
 problem. To have more numerical precision in the comparison in our experiments,
 we calculate the Area Under the CDF Curve (AUC), using only the NRMSE range
 [0,0.5] with a step of $10^{-3}$.

 \subsection{General training setup}
 
To implement our framework, we use:
 \begin{inparaitem}[-]
\item a DNN with four layers $K=4$ for the main task;
\item an input AE with one encoding layer $K_{in}=1$ and one decoding layer;
\item an output AE with one encoding layer and one decoding layer $K_{out}=1$.
 \end{inparaitem}
Referring to Fig.\ref{fig:fig0.1}, the size of the input representation $\x$ and estimation $\xest$ is $2500=50 \times 50$;
the size of the output representation $\y$ and estimation $\yest$ is $136= 68 \times 2$, given the 68 landmarks in a 2D plane;
the dimension of intermediate spaces $\xproj$ and $\yproj$ have been set to 1025 and 64 respectively;
finally, the hidden layer in the $m$ link between  $\xproj$ and $\yproj$ is composed of 512 units.
The size of each layer has been set using a validation procedure on the LFPW validation set.

Sigmoid activation functions are used everywhere in the main NN and in the two AEs, except for the last layer of the main NN and the tied last layer of output AE which use a hyperbolic tangent activation function to suite the range $[-1, \;1]$ for the output $\y$.
 
 We use the same architecture through all the experiments for the different
 training configurations. To distinguish between the multiple configurations we
 set the following notations:
\begin{enumerate}
\item \textbf{MLP}, a DNN for the main task with no concomitant training;
\item \textbf{MLP + in}, a DNN with input AE parallel training;
\item \textbf{MLP + out}, a DNN with output AE parallel training;
\item \textbf{MLP + in + out}, a DNN with both input and output reconstruction secondary tasks.
\end{enumerate}
We recall that the auto-encoders are used only during the training phase. In the test phase, they are dropped. Therefore, the final test networks have the same architecture in all the different configurations.

Beside these configurations, we consider the mean shape (the average of the $\y$ in the training data) as a simple predictive model. For each test image, we predict the same estimated mean shape over the train set.

To clarify the benefit of our approach, all the configurations must start from the same initial weights to make sure that the obtained improvement is due to the training algorithm, not to the random initialization.

For the input reconstruction tasks, we use a denoising auto-encoder with a
corruption level of $20\%$ for the first hidden layer. For the output
reconstruction task, we use a simple auto-encoder. To avoid overfitting, the auto-encoders are trained using $L_2$ regularization with a weight decay of $10^{-2}$.

In all the configurations, the update of the parameters of each task (supervised and unsupervised) is performed using Stochastic Gradient Descent with momentum
\cite{sutskever1} with a constant momentum coefficient of $0.9$. We use
mini-batch size of 10. The training is performed for 1000 epochs with a learning rate of $10^{-3}$.

\begin{figure}[!htb]
   \centering
   \includegraphics[scale=0.5]{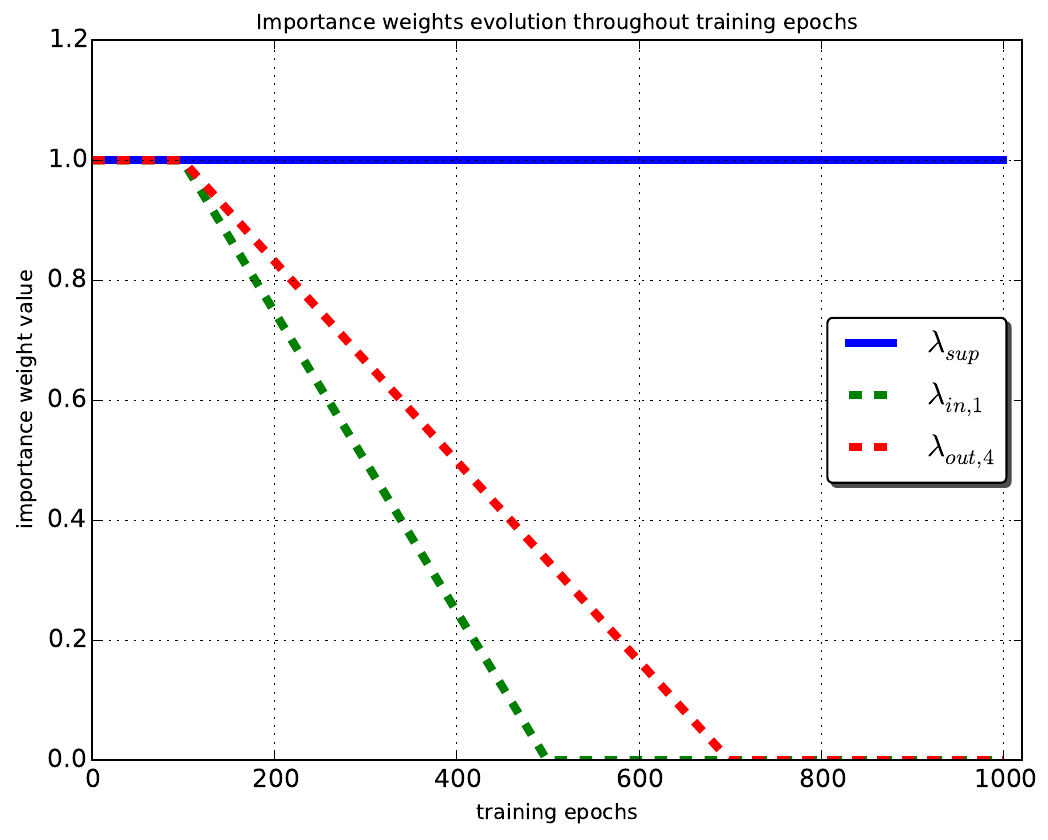}
   \caption{Linear evolution of the importance weights during training.}
   \label{fig:fig000}
 \end{figure}

In these experiments, we propose to use a simple linear evolution scheme for the importance weights $\lambda_{sup}$ (supervised task), $\lambda_{in}$ (input task) and  $\lambda_{out}$ (output task). We retain the evolution proposed in \cite{bel16},  and presented in Fig.\ref{fig:fig000}.

The hyper-parameters (learning rate, batch size, momentum coefficient, weight decay, the importance weights) have been optimized on the LFPW validation set. We apply the same optimized hyper-parameters for HELEN dataset.

Using these configurations, we perform two types of experiments: with and
without unlabeled data. We present in the next sections the obtained results.

\subsubsection{Experiments with fully labeled data}
\label{sssec:naug}
In this setup, we use the provided labeled data from each set in a classical way. For LFPW set, we use the 676 available samples for training and 135 samples for validation. For HELEN set, we use 1800 samples for training and 200 samples for validation. 

In order to evaluate the different configurations, we first calculate the Mean Squared Error (MSE) of the best models found using the validation during the training. Column 1 (no unlabeled data) of Tab.\ref{tab:tab1}, \ref{tab:tab2} shows the MSE over the train and valid sets of LFPW and HELEN datasets, respectively. Compared to an MLP alone, adding the input training of the first hidden layer slightly reduces the train and validation error in both datasets. Training the output layer also reduces the train and validation error, with a more important factor. Combining the input train of the first hidden layer and output train of the last layer gives the best performance. We plot the tracked MSE over the train and valid sets of HELEN dataset in Fig.\ref{fig:fig4:fig2}, \ref{fig:fig4:fig3}. One can see that the input training reduces slightly the validation MSE. The output training has a major impact over the training speed and the generalization of the model which suggests that output training is useful in the case of structured output problems. Combining the input and the output training improves even more the generalization. Similar behavior was found on LFPW dataset.

At a second time, we evaluate each configuration over the test set of each datasets using the $CDF_{0.1}$ metric. The results are depicted in Tab.\ref{tab:tab3}, \ref{tab:tab4} in the first column for LFPW and HELEN datasets, respectively. Similarly to the results  previously found over the train and validation set, one can see that the joint training (supervised, input, output) outperforms all the other configurations in terms of $CDF_{0.1}$ and AUC. The CDF curves in Fig.\ref{fig:fig4} also confirms this result. Compared to the global DNN in \cite{zhang14} over LFPW test set, our joint trained MLP performs better (\cite{zhang14}: $CDF_{0.1}=65\%$, ours: $CDF_{0.1}=69.64\%$), despite the fact that their model was trained using larger supervised dataset (combination of multiple supervised datasets beside LFPW).

\begin{figure}[!htb]
   \centering
   \includegraphics[width=30mm,height=30mm]{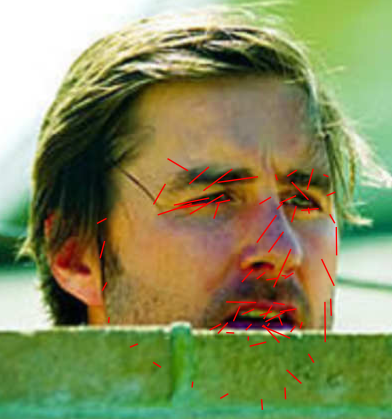}
   \includegraphics[width=30mm,height=30mm]{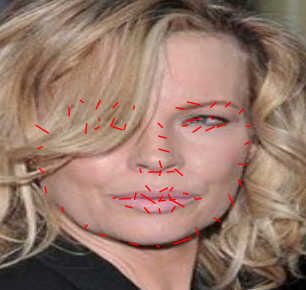}
   \includegraphics[width=30mm,height=30mm]{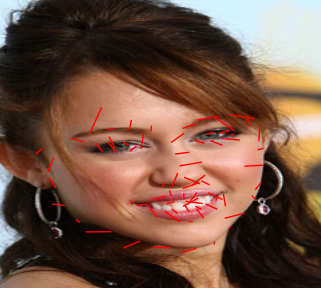} \\
   \includegraphics[width=30mm,height=30mm]{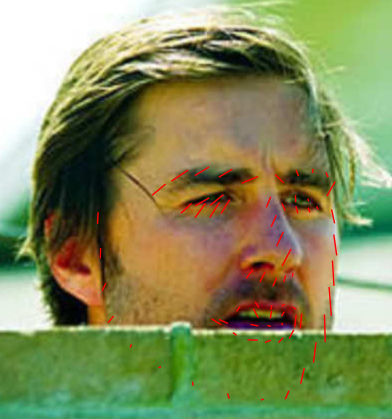}
   \includegraphics[width=30mm,height=30mm]{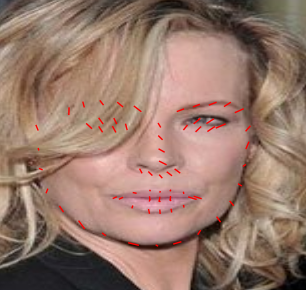}
   \includegraphics[width=30mm,height=30mm]{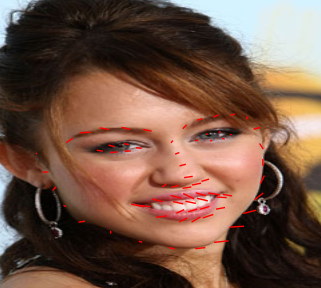}
   \caption{Examples of prediction on LFPW test set. For visualizing errors, red segments have been drawn between ground truth and predicted landmark. Top row: MLP. Bottom row: MLP+in+out. (no unlabeled data)}
   \label{fig:figlfpw}
   \end{figure}

\begin{figure}[!htb]
   \centering
   \includegraphics[width=30mm,height=30mm]{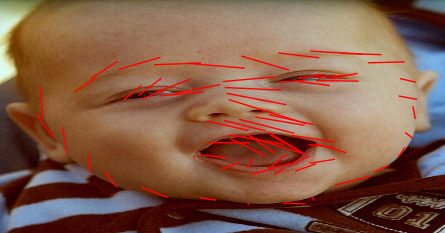}
   \includegraphics[width=30mm,height=30mm]{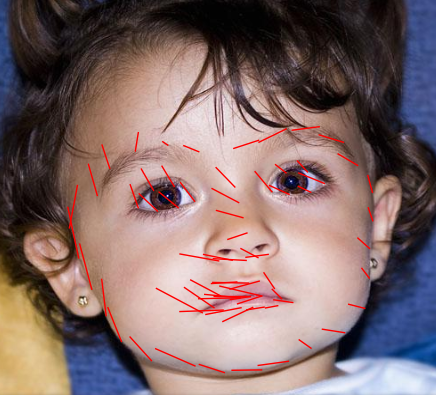}
   \includegraphics[width=30mm,height=30mm]{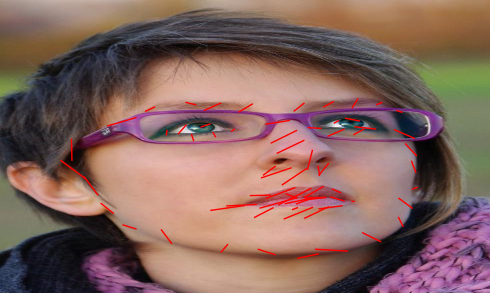} \\
   \includegraphics[width=30mm,height=30mm]{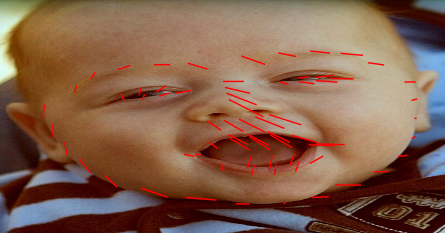}
   \includegraphics[width=30mm,height=30mm]{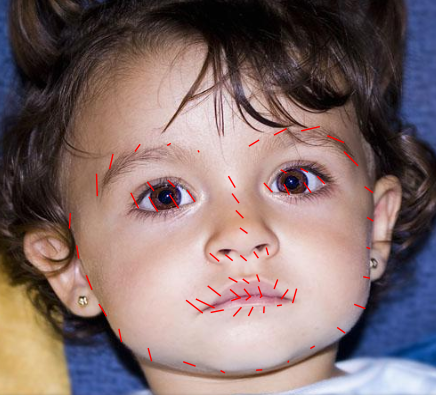}
   \includegraphics[width=30mm,height=30mm]{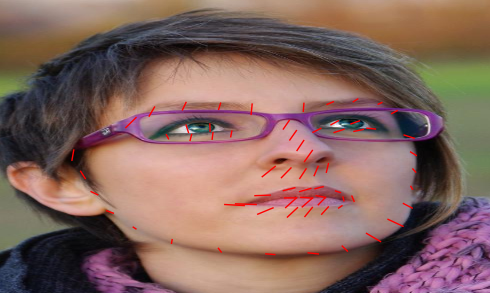}
   \caption{Examples of prediction on HELEN test set. Top row: MLP. Bottom row: MLP+in+out. (no unlabeled data)}
   \label{fig:fighelen}
   \end{figure}

\begin{figure}[!htb]
   \centering
   \subfigure[]{
   \includegraphics[scale=0.65]{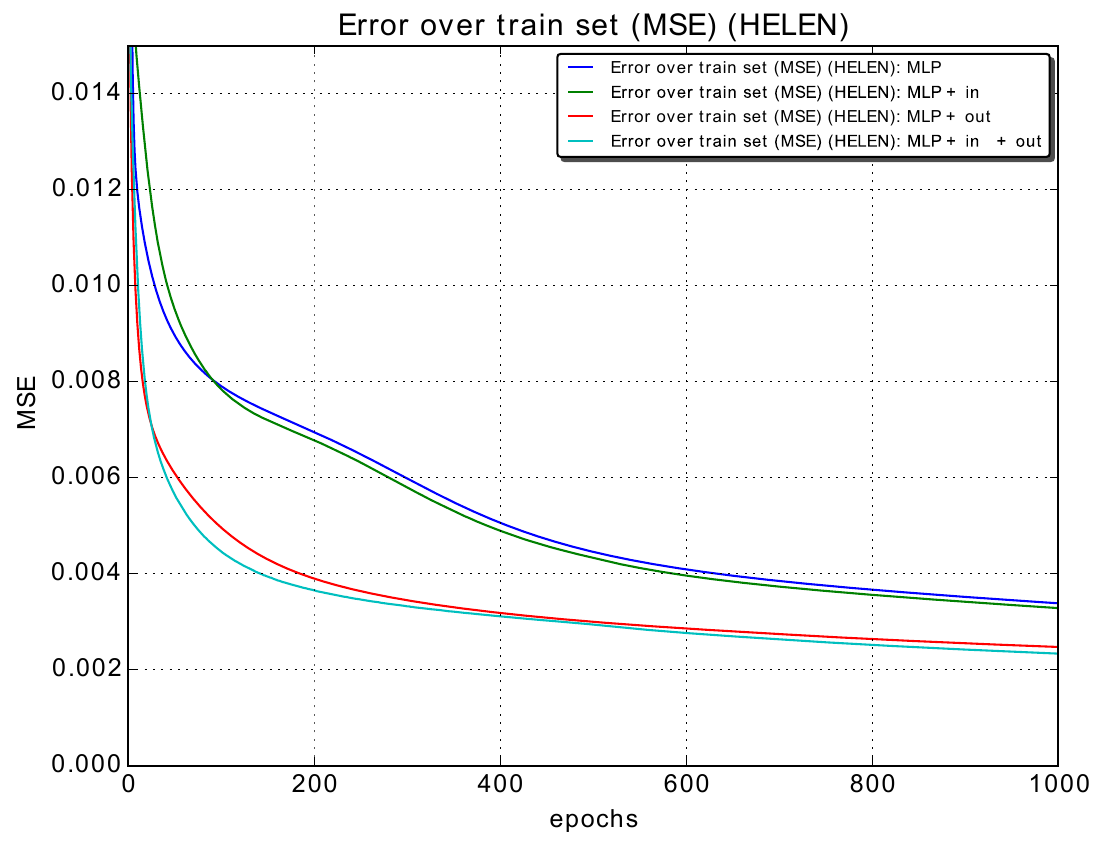}
   \label{fig:fig4:fig2}
   }%
   \\
   ~
   \subfigure[]{
   \includegraphics[scale=0.65]{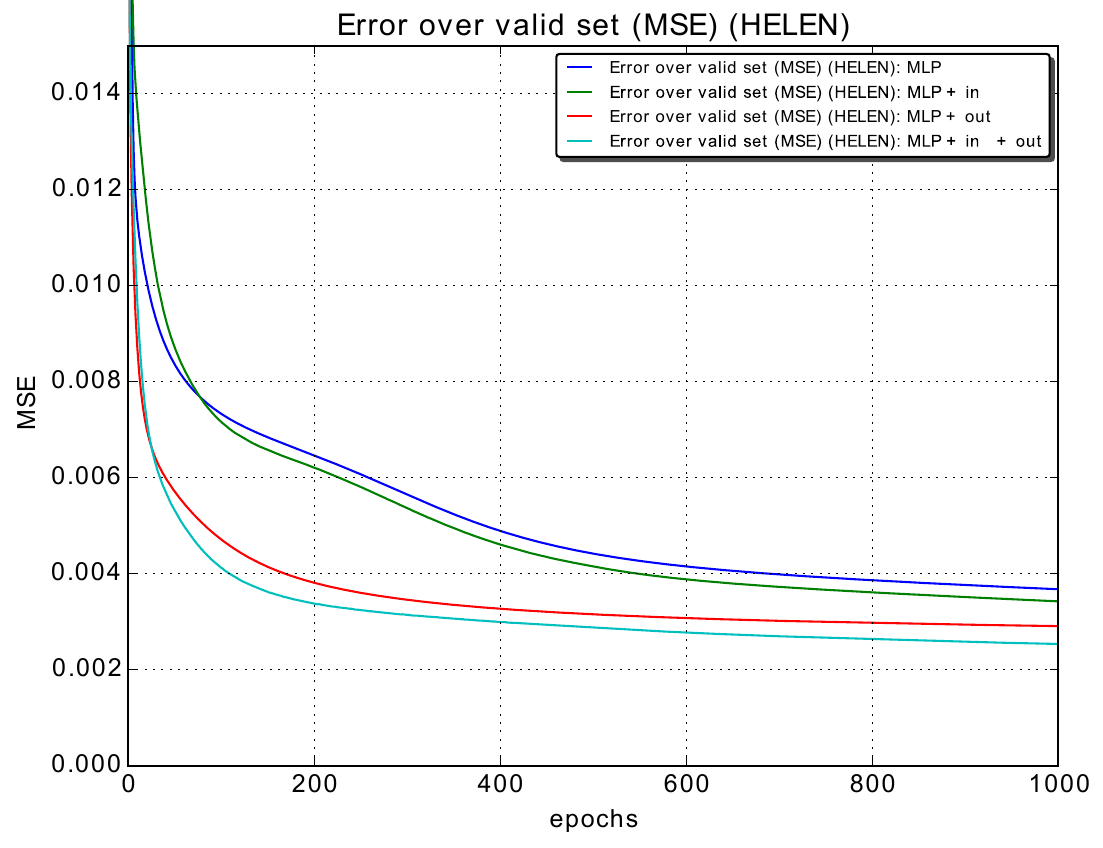}
   \label{fig:fig4:fig3}
   }
   \caption{MSE during training epochs over HELEN train (a) and valid (b)
     sets using different training setups for the MLP.}
   \label{fig:fig5}
 \end{figure}

\begin{figure}[!htb]
   \centering
   \subfigure[]{
   \includegraphics[scale=0.63]{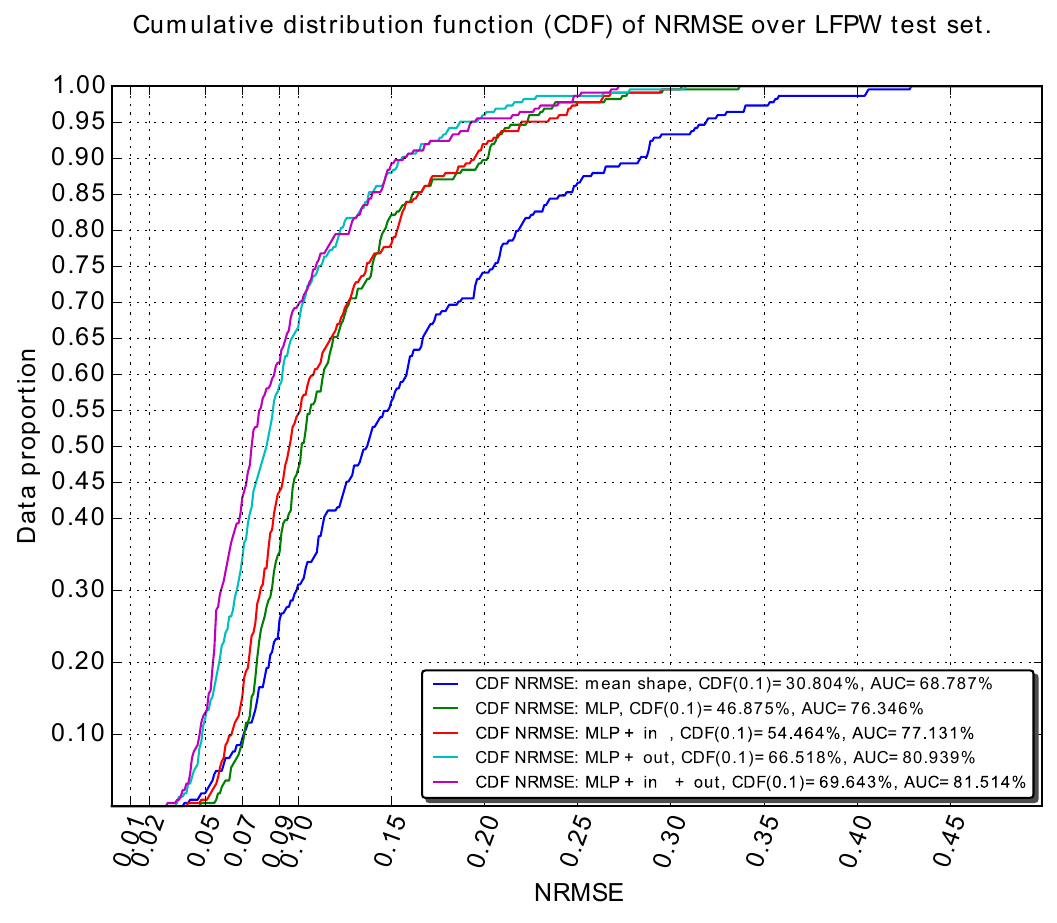}
   \label{fig:fig4:fig0}
   }%
   \\
   ~
   \subfigure[]{
   \includegraphics[scale=0.63]{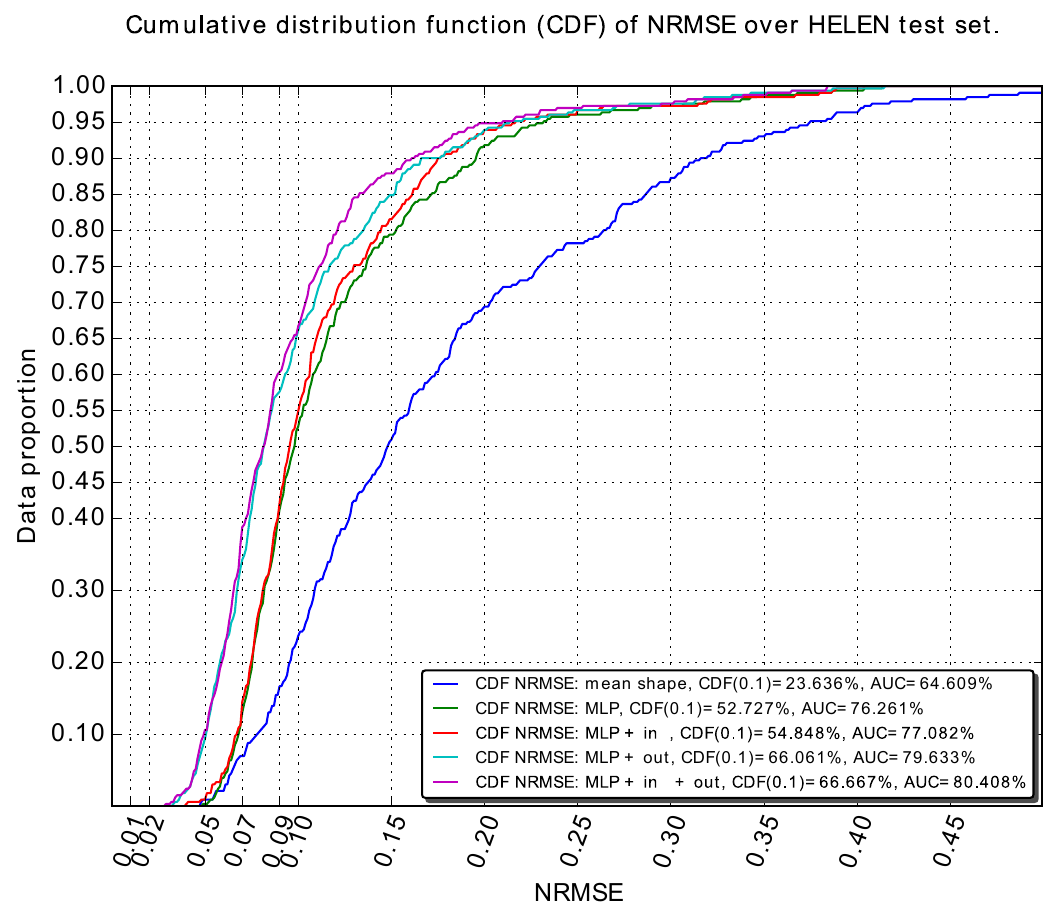}
   \label{fig:fig4:fig1}
   }
   \caption{CDF curves of different configurations on: (a) LFPW, (b) HELEN.}
   \label{fig:fig4}
 \end{figure}

An illustrative result of our method is presented in Fig.\ref{fig:figlfpw}, \ref{fig:fighelen} for LFPW and HELEN using an MLP and MLP with input and output training.

  \begin{table}[!htb]
   \centering
   \caption{MSE over LFPW: train and valid sets, at the end of training with and
     without unlabeled data.}
   \label{tab:tab1}
   \resizebox{0.9\textwidth}{!}{
	\sisetup{detect-weight=true,detect-inline-weight=math}
 	\begin{tabular}{l|c|c||c|c|}
 		\cline{2-5}
 		&\multicolumn{2}{|c||}{\textbf{No unlabeled data}} 
 &\multicolumn{2}{|c|}{\textbf{With unlabeled data}}\\
 		\cline{2-5}
 		&\multicolumn{1}{|c|}{MSE train}&\multicolumn{1}{|c||}{MSE valid}&\multicolumn{1}{|c|}{MSE train}&\multicolumn{1}{|c|}{MSE valid}\\
                \hline
 		\multicolumn{1}{ |l | } {\textbf{Mean 
                    shape}}&
                \num{7.74d-3} &\num{8.07d-3}&\num{7.78d-3}&\num{8.14d-3}\\
 		\hline
 		\hline
 		\multicolumn{1}{ |l | } {\textbf{MLP}}&
                \num{3.96d-3}&\num{4.28d-3}&-&-\\
 		\hline
                \hline
 		\multicolumn{1}{ |l | } {\textbf{MLP + in}}&
                \num{3.64d-3}&\num{3.80d-3}&\num{1.44d-3}&\num{2.62d-3}\\
 		\hline
 		\multicolumn{1}{ |l | } {\textbf{MLP + out}}&
                \num{2.31d-3}&\num{2.99d-3}&\num{1.51d-3}&\num{2.79d-3}\\
 		\hline
 		\multicolumn{1}{ |l | } {\textbf{MLP + in + out}}&
                \textbf{\num{2.12d-3}}&\textbf{\num{2.56d-3}}&\textbf{\num{1.10d-3}}&\textbf{\num{2.23d-3}}\\
 		\hline
 	\end{tabular}
	}
\end{table}

   \begin{table}[!htb]
   \centering
   \caption{MSE over HELEN: train and valid sets, at the end of training with and
     without data augmentation.}
   \label{tab:tab2}
   \resizebox{0.9\textwidth}{!}{
   	\sisetup{detect-weight=true,detect-inline-weight=math}
 	\begin{tabular}{l|c|c||c|c|}
 		\cline{2-5}
 		&\multicolumn{2}{|c||}{\textbf{Fully labeled data only }} 
 &\multicolumn{2}{|c|}{\textbf{Adding unlabeled or label-only data}}\\
 		\cline{2-5}
 		&\multicolumn{1}{|c|}{MSE train}&\multicolumn{1}{|c||}{MSE valid}&\multicolumn{1}{|c|}{MSE train}&\multicolumn{1}{|c|}{MSE valid}\\
                \hline
 		\multicolumn{1}{ |l | } {\textbf{Mean 
                    shape}}&
                \num{7.59d-3}&\num{6.95d-3}&\num{7.60d-3}&\num{.95d-3}\\
 		\hline
 		\hline
 		\multicolumn{1}{ |l | } {\textbf{MLP}}&
                \num{3.39d-3}&\num{3.67d-3}&-&-\\
 		\hline
                \hline
 		\multicolumn{1}{ |l | } {\textbf{MLP + in}}&
                \num{3.28d-3}&\num{3.42d-3}&\num{2.31d-3}&\num{2.81d-3}\\
 		\hline
 		\multicolumn{1}{ |l | } {\textbf{MLP + out}}&
                \num{2.48d-3}&\num{2.90d-3}&\num{2.00d-3}&\num{2.74d-3}\\
 		\hline
 		\multicolumn{1}{ |l | } {\textbf{MLP + in + out}}&
               \textbf{\num{2.34d-3}}&\textbf{\num{2.53d-3}}&\textbf{\num{1.92d-3}}&\textbf{\num{2.40d-3}}\\
  		\hline
 	\end{tabular}
        }
 \end{table}


\begin{table}[!htb]
   \centering
   \caption{\textbf{AUC} and $\mathbf{CDF_{0.1}}$ performance over LFPW test dataset with and without unlabeled data.}
   \label{tab:tab3}
   \resizebox{.8\textwidth}{!}{
 	\begin{tabular}{c|c|c||c|c|}
 		\cline{2-5}
 		&\multicolumn{2}{|c||}{\textbf{Fully labeled data only}} 
 &\multicolumn{2}{|c|}{\textbf{Adding unlabeled or label-only data}}\\
 		\cline{2-5}
 		&\textbf{AUC}&$\mathbf{CDF_{0.1}}$&\textbf{AUC}&$\mathbf{CDF_{0.1}}$\\
                \hline
 		\multicolumn{1}{ |l | } {\textbf{Mean 
                    shape}}&
                68.78\%&30.80\%&77.81\%&22.33\%\\
 		\hline
 		\hline
 		\multicolumn{1}{ |l | } {\textbf{MLP}}&
                76.34\%&46.87\%&-&-\\
 		\hline
                \hline
 		\multicolumn{1}{ |l | } {\textbf{MLP + in}}&
                77.13\%&54.46\%&80.78\%&67.85\%\\
 		\hline
 		\multicolumn{1}{ |l | } {\textbf{MLP + out}}&
                80.93\%&66.51\%&81.77\%&67.85\%\\
 		\hline
 		\multicolumn{1}{ |l | } {\textbf{MLP + in + out}}&
                \textbf{81.51\%}&\textbf{69.64\%}&\textbf{82.48\%}&\textbf{71.87\%}\\
 		\hline
 	\end{tabular}
     }
 \end{table}

\begin{table}[!htb]
   \centering
   \caption{\textbf{AUC} and $\mathbf{CDF_{0.1}}$ performance over HELEN test dataset with and without unlabeled data.}
   \label{tab:tab4}
   \resizebox{.8\textwidth}{!}{
 	\begin{tabular}{c|c|c||c|c|}
 		\cline{2-5}
 		&\multicolumn{2}{|c||}{\textbf{Fully labeled data only}} 
 &\multicolumn{2}{|c|}{\textbf{Adding unlabeled or label-only data}}\\
 		\cline{2-5}
 		&\textbf{AUC}&$\mathbf{CDF_{0.1}}$&\textbf{AUC}&$\mathbf{CDF_{0.1}}$\\
                \hline
 		\multicolumn{1}{ |l | } {\textbf{Mean 
                    shape}}&
                64.60\%&23.63\%&64.76\%&23.23\%\\
 		\hline
 		\hline
 		\multicolumn{1}{ |l | } {\textbf{MLP}}&
                76.26\%&52.72\%&-&-\\
 		\hline
                \hline
 		\multicolumn{1}{ |l | } {\textbf{MLP + in}}&
                77.08\%&54.84\%&79.25\%&63.33\%\\
                \hline
 		\multicolumn{1}{ |l | } {\textbf{MLP + out}}&
                79.63\%&66.60\%&80.48\%&65.15\%\\
 		\hline
 		\multicolumn{1}{ |l | } {\textbf{MLP + in + out}}&
                \textbf{80.40\%}&\textbf{66.66\%}&\textbf{81.27\%}&\textbf{71.51\%}\\
  		\hline
 	\end{tabular}
      }
 \end{table}

 \subsubsection{Data augmentation using unlabeled data or label-only data}
 In this section, we experiment our approach when adding unlabeled data (input and output). Unlabeled data (i.e. image faces without the landmarks annotation) are abundant and can be found easily for example from other datasets or from the Internet which makes it practical and realistic. In our case, we use image faces from another dataset.

In the other hand, label-only data (i.e. the landmarks annotation without image faces) are more difficult to obtain because we usually have the annotation based on the image faces. One way to obtain accurate and realistic facial landmarks without image faces is to use a 3D face model as a generator. We use an easier way to obtain facial landmarks annotation by taking them from another dataset.

In this experiment, in order to add unlabeled data for LFPW dataset, we take all the image faces of HELEN dataset (train, valid and test) and vice versa for HELEN dataset by taking all LFPW image faces as unlabeled data. The same experiment is performed for the label-only data using the facial landmarks annotation. We summarize the size of each train set in Tab.\ref{tab:tab5}..

\begin{table}[!htb]
\centering
\caption{Size of augmented LFPW and HELEN train sets.}
\label{tab:tab5}
\resizebox{\textwidth}{!}{
  \begin{tabular}{|c|c|c|c|}
    \hline
    Train set / size of& Supervised data& Unsupervised input $\x$ & Unsupervised
    output $\y$\\
    \hline
    LFPW&676&2330&2330\\
    \hline
    HELEN&1800&1035&1035\\
    \hline
\end{tabular}
}
\end{table}
We use the same validation sets as in Sec.\ref{sssec:naug} in order to have a fair comparison. The MSE are presented in the second column of Tab.\ref{tab:tab1}, \ref{tab:tab2} over LFPW and HELEN datasets. One can see that adding unlabeled data decreases the MSE over the train and validation sets. Similarly, we found that the input training along with the output training gives the best results. Identically,
these results are translated in terms of $CDF_{0.1}$ and AUC over the test sets (Tab.\ref{tab:tab3}, \ref{tab:tab4}). All these results suggest that adding unlabeled input and output data can improve the generalization of our framework and the training speed.

%% file: 6-conclusion.tex
\section{Conclusion and Future Work}
\label{sec:conclusion}
In this paper, we tackled structured output prediction problems as a representation learning problem. We have proposed a generic multi-task training framework as a regularization scheme for structured output prediction models. It has been instantiated through a deep neural network model which learns the input and output distributions using auto-encoders while learning the supervised task $\x \to \y$. Moreover, we explored the possibility of using the output labels $\y$ without their corresponding input data $\x$ which showed more improvement in the generalization. Using a parallel scheme allows an interaction between the main supervised task and the unsupervised tasks which helped preventing the overfitting of the main task.

We evaluated our training method on a facial landmark detection task over two public datasets. The obtained results showed that our proposed regularization scheme improves the generalization of neural networks model and speeds up their training. We believe that our approach provides an alternative for training deep architectures for structured output prediction where it allows the use of unlabeled input and label of the output data. 

As a future work, we plan to evolve automatically the importance weights of the tasks. For that and in order to better guide their evolution, we can consider the use of different indicators based on the training and the validation errors instead of the learning epochs only. Furthermore, one may consider other kind of models instead of simple auto-encoders in order to learn the output distribution. More specifically, generative models such as variational and adversarial auto-encoders \cite{DBLP:journals/corr/MakhzaniSJG15} could be explored.

%% file: main.bbl
\begin{thebibliography}{10}

\bibitem{auliGQZ13}
Michael Auli, Michel Galley, Chris Quirk, and Geoffrey Zweig.
\newblock Joint language and translation modeling with recurrent neural
  networks.
\newblock In {\em Proceedings of the 2013 Conference on Empirical Methods in
  Natural Language Processing, {EMNLP} 2013, 18-21 October 2013, Grand Hyatt
  Seattle, Seattle, Washington, USA, {A} meeting of SIGDAT, a Special Interest
  Group of the {ACL}}, pages 1044--1054, 2013.

\bibitem{belangerM16}
David Belanger and Andrew McCallum.
\newblock Structured prediction energy networks.
\newblock In {\em Proceedings of the 33nd International Conference on Machine
  Learning, {ICML} 2016, New York City, NY, USA, June 19-24, 2016}, pages
  983--992, 2016.

\bibitem{bel16}
S.~Belharbi, R.Hérault, C.~Chatelain, and S.~Adam.
\newblock Deep multi-task learning with evolving weights.
\newblock In {\em European Symposium on Artificial Neural Networks ({ESANN})},
  2016.

\bibitem{belhumeur11}
Peter~N. Belhumeur, David~W. Jacobs, David~J. Kriegman, and Neeraj Kumar.
\newblock {Localizing parts of faces using a consensus of exemplars.}
\newblock In {\em {CVPR}}, pages 545--552. IEEE, 2011.

\bibitem{bengio07}
Yoshua Bengio, Pascal Lamblin, Dan Popovici, and Hugo Larochelle.
\newblock {Greedy Layer-Wise Training of Deep Networks}.
\newblock In B.~Sch{\"o}lkopf, J.C. Platt, and T.~Hoffman, editors, {\em
  {NIPS}}, pages 153--160. 2007.

\bibitem{bikel99}
Daniel~M Bikel, Richard Schwartz, and Ralph~M Weischedel.
\newblock {An algorithm that learns what's in a name}.
\newblock {\em Machine learning}, 34(1-3):211--231, 1999.

\bibitem{caruana97ML}
R.~Caruana.
\newblock Multitask learning.
\newblock {\em Machine Learning}, 28(1):41--75, 1997.

\bibitem{ciresanGGS12}
Dan~C. Ciresan, Alessandro Giusti, Luca~Maria Gambardella, and J{\"{u}}rgen
  Schmidhuber.
\newblock Deep neural networks segment neuronal membranes in electron
  microscopy images.
\newblock In {\em Advances in Neural Information Processing Systems 25: 26th
  Annual Conference on Neural Information Processing Systems 2012. Proceedings
  of a meeting held December 3-6, 2012, Lake Tahoe, Nevada, United States.},
  pages 2852--2860, 2012.

\bibitem{collobert08ICML}
R.~Collobert and J.~Weston.
\newblock A unified architecture for natural language processing: deep neural
  networks with multitask learning.
\newblock In {\em ICML}, pages 160--167, 2008.

\bibitem{cristinacce06}
D.~Cristinacce and T.~Cootes.
\newblock {Feature Detection and Tracking with Constrained Local Models}.
\newblock In {\em {BMVC}}, pages 95.1--95.10, 2006.

\bibitem{elyacoubi02}
M.~El-Yacoubi, M.~Gilloux, and J-M Bertille.
\newblock {A statistical approach for phrase location and recognition within a
  text line: An application to street name recognition}.
\newblock {\em IEEE PAMI}, 24(2):172--188, 2002.

\bibitem{farabet13}
Cl{\'e}ment Farabet, Camille Couprie, Laurent Najman, and Yann LeCun.
\newblock {Learning Hierarchical Features for Scene Labeling}.
\newblock {\em {IEEE} PAMI}, 35(8):1915--1929, 2013.

\bibitem{fridman93}
Moshe Fridman.
\newblock {\em {Hidden markov model regression}}.
\newblock PhD thesis, Graduate School of Arts and Sciences, University of
  Pennsylvania, 1993.

\bibitem{gravesJ14}
Alex Graves and Navdeep Jaitly.
\newblock Towards end-to-end speech recognition with recurrent neural networks.
\newblock In {\em Proceedings of the 31th International Conference on Machine
  Learning, {ICML} 2014, Beijing, China, 21-26 June 2014}, pages 1764--1772,
  2014.

\bibitem{graves2009novel}
Alex Graves, Marcus Liwicki, Santiago Fern{\'a}ndez, Roman Bertolami, Horst
  Bunke, and J{\"u}rgen Schmidhuber.
\newblock A novel connectionist system for unconstrained handwriting
  recognition.
\newblock {\em IEEE transactions on pattern analysis and machine intelligence},
  31(5):855--868, 2009.

\bibitem{HoffmanWYD16}
Judy Hoffman, Dequan Wang, Fisher Yu, and Trevor Darrell.
\newblock Fcns in the wild: Pixel-level adversarial and constraint-based
  adaptation.
\newblock {\em CoRR}, abs/1612.02649, 2016.

\bibitem{JaderbergSVZ14b2014}
Max Jaderberg, Karen Simonyan, Andrea Vedaldi, and Andrew Zisserman.
\newblock Deep structured output learning for unconstrained text recognition.
\newblock {\em CoRR}, abs/1412.5903, 2014.

\bibitem{jones99}
David~T. Jones.
\newblock {Protein secondary structure prediction based on position-specific
  scoring matrices}.
\newblock {\em Journal of Molecular Biology}, 292(2):195--202, 1999.

\bibitem{karpathyL15}
Andrej Karpathy and Fei{-}Fei Li.
\newblock Deep visual-semantic alignments for generating image descriptions.
\newblock In {\em {IEEE} Conference on Computer Vision and Pattern Recognition,
  {CVPR} 2015, Boston, MA, USA, June 7-12, 2015}, pages 3128--3137, 2015.

\bibitem{krizhevsky12}
Alex Krizhevsky, Ilya Sutskever, and Geoffrey~E. Hinton.
\newblock {ImageNet Classification with Deep Convolutional Neural Networks}.
\newblock In F.~Pereira, C.J.C. Burges, L.~Bottou, and K.Q. Weinberger,
  editors, {\em {Advances in Neural Information Processing Systems 25}}, pages
  1097--1105. Curran Associates, Inc., 2012.

\bibitem{lafferty01}
John~D. Lafferty, Andrew McCallum, and Fernando C.~N. Pereira.
\newblock {Conditional Random Fields: Probabilistic Models for Segmenting and
  Labeling Sequence Data}.
\newblock In {\em { ICML}}, pages 282--289, 2001.

\bibitem{le12}
Vuong Le, Jonathan Brandt, Zhe Lin, Lubomir~D. Bourdev, and Thomas~S. Huang.
\newblock {Interactive Facial Feature Localization}.
\newblock In {\em {ECCV, 2012, Proceedings, Part {III}}}, pages 679--692, 2012.

\bibitem{lerouge15}
J.~Lerouge, R.~Herault, C.~Chatelain, F.~Jardin, and R.~Modzelewski.
\newblock {IODA: An Input Output Deep Architecture for image labeling}.
\newblock {\em Pattern Recognition}, 2015.

\bibitem{LiUBBSB16}
Xirong Li, Tiberio Uricchio, Lamberto Ballan, Marco Bertini, Cees G.~M. Snoek,
  and Alberto~Del Bimbo.
\newblock Socializing the semantic gap: {A} comparative survey on image tag
  assignment, refinement, and retrieval.
\newblock {\em {ACM} Comput. Surv.}, 49(1):14:1--14:39, 2016.

\bibitem{liu2014}
Shujie Liu, Nan Yang, Mu~Li, and Ming Zhou.
\newblock A recursive recurrent neural network for statistical machine
  translation.
\newblock In {\em Proceedings of the 52nd Annual Meeting of the Association for
  Computational Linguistics (Volume 1: Long Papers)}, pages 1491--1500,
  Baltimore, Maryland, June 2014. Association for Computational Linguistics.

\bibitem{longSD15}
Jonathan Long, Evan Shelhamer, and Trevor Darrell.
\newblock Fully convolutional networks for semantic segmentation.
\newblock In {\em {IEEE} Conference on Computer Vision and Pattern Recognition,
  {CVPR} 2015, Boston, MA, USA, June 7-12, 2015}, pages 3431--3440, 2015.

\bibitem{DBLP:journals/corr/MakhzaniSJG15}
Alireza Makhzani, Jonathon Shlens, Navdeep Jaitly, and Ian~J. Goodfellow.
\newblock Adversarial autoencoders.
\newblock {\em CoRR}, abs/1511.05644, 2015.

\bibitem{mnihLH11}
Volodymyr Mnih, Hugo Larochelle, and Geoffrey~E. Hinton.
\newblock Conditional restricted boltzmann machines for structured output
  prediction.
\newblock In {\em {UAI} 2011, Proceedings of the Twenty-Seventh Conference on
  Uncertainty in Artificial Intelligence, Barcelona, Spain, July 14-17, 2011},
  pages 514--522, 2011.

\bibitem{nicolas06}
St{\'e}phane Nicolas, Thierry Paquet, and Laurent Heutte.
\newblock {A Markovian Approach for Handwritten Document Segmentation.}
\newblock In {\em {ICPR (3)}}, pages 292--295, 2006.

\bibitem{ningDLPBB05}
F.~Ning, D.~Delhomme, Yann LeCun, F.~Piano, L{\'{e}}on Bottou, and Paolo~Emilio
  Barbano.
\newblock Toward automatic phenotyping of developing embryos from videos.
\newblock {\em {IEEE} Trans. Image Processing}, 14(9):1360--1371, 2005.

\bibitem{nohHH15}
Hyeonwoo Noh, Seunghoon Hong, and Bohyung Han.
\newblock Learning deconvolution network for semantic segmentation.
\newblock In {\em 2015 {IEEE} International Conference on Computer Vision,
  {ICCV} 2015, Santiago, Chile, December 7-13, 2015}, pages 1520--1528, 2015.

\bibitem{noto12}
Keith Noto and Mark Craven.
\newblock {Learning Hidden Markov Models for Regression using Path
  Aggregation}.
\newblock {\em CoRR}, abs/1206.3275, 2012.

\bibitem{och03}
Franz~Josef Och.
\newblock {Minimum error rate training in statistical machine translation}.
\newblock In {\em {Proceedings of the ACL}}, volume~1, 2003.

\bibitem{rabiner89}
Lawrence Rabiner.
\newblock {A tutorial on hidden Markov models and selected applications in
  speech recognition}.
\newblock {\em Proceedings of the IEEE}, 77(2):257--286, 1989.

\bibitem{ronnebergerFB15}
Olaf Ronneberger, Philipp Fischer, and Thomas Brox.
\newblock U-net: Convolutional networks for biomedical image segmentation.
\newblock In {\em Medical Image Computing and Computer-Assisted Intervention -
  {MICCAI} 2015 - 18th International Conference Munich, Germany, October 5 - 9,
  2015, Proceedings, Part {III}}, pages 234--241, 2015.

\bibitem{sagonas13}
C.~Sagonas, G.~Tzimiropoulos, S.~Zafeiriou, and M.~Pantic.
\newblock A semi-automatic methodology for facial landmark annotation.
\newblock In {\em CVPR Workshops}, pages 896--903, 2013.

\bibitem{schmid94}
H.~Schmid.
\newblock {Part-of-speech tagging with neural networks}.
\newblock {\em conference on Computational linguistics}, 12:44--49, 1994.

\bibitem{sleator95}
Daniel~Dominic Sleator and David Temperley.
\newblock {Parsing English with a Link Grammar}.
\newblock {\em CoRR}, 1995.

\bibitem{SohnLY15}
Kihyuk Sohn, Honglak Lee, and Xinchen Yan.
\newblock Learning structured output representation using deep conditional
  generative models.
\newblock In {\em NIPS 2015}, pages 3483--3491, 2015.

\bibitem{Stuner16}
Bruno Stuner, Cl{\'{e}}ment Chatelain, and Thierry Paquet.
\newblock Cohort of {LSTM} and lexicon verification for handwriting recognition
  with gigantic lexicon.
\newblock {\em CoRR}, abs/1612.07528, 2016.

\bibitem{sutskever1}
I.~Sutskever, J.~Martens, G.~Dahl, and G.~Hinton.
\newblock {On the importance of initialization and momentum in deep learning}.
\newblock In {\em {ICML}}, volume~28, pages 1139--1147, 2013.

\bibitem{sutskeverVL14}
Ilya Sutskever, Oriol Vinyals, and Quoc~V. Le.
\newblock Sequence to sequence learning with neural networks.
\newblock In {\em Advances in Neural Information Processing Systems 27: Annual
  Conference on Neural Information Processing Systems 2014, December 8-13 2014,
  Montreal, Quebec, Canada}, pages 3104--3112, 2014.

\bibitem{syed09}
U.~Syed and G.~Yona.
\newblock Enzyme function prediction with interpretable models.
\newblock {\em Computational Systems Biology. Humana press}, pages 373--420,
  2009.

\bibitem{szummer04}
M.~Szummer and Y.~Qi.
\newblock {Contextual Recognition of Hand-drawn Diagrams with Conditional
  Random Fields}.
\newblock In {\em {IWFHR}}, pages 32--37, 2004.

\bibitem{tsechpenakis07}
G.~Tsechpenakis, Jianhua Wang, B.~Mayer, and D.~Metaxas.
\newblock {Coupling CRFs and Deformable Models for 3D Medical Image
  Segmentation}.
\newblock In {\em {ICCV}}, pages 1--8, 2007.

\bibitem{vincent10}
P.~Vincent, H.~Larochelle, I.~Lajoie, Y.~Bengio, and P.~Manzagol.
\newblock {Stacked Denoising Autoencoders: Learning Useful Representations in a
  Deep Network with a Local Denoising Criterion}.
\newblock {\em JMLR}, 11:3371--3408, 2010.

\bibitem{zen09}
H.~Zen, K.~Tokuda, and A.~Black.
\newblock {Statistical parametric speech synthesis}.
\newblock {\em Speech Communication}, 51(11):1039--1064, 2009.

\bibitem{zhang14}
J.~Zhang, S.~Shan, M.~Kan, and X.~Chen.
\newblock {Coarse-to-Fine Auto-Encoder Networks {(CFAN)} for Real-Time Face
  Alignment}.
\newblock In {\em {ECCV, Part {II}}}, pages 1--16, 2014.

\bibitem{ZhangDG17}
Yang Zhang, Philip David, and Boqing Gong.
\newblock Curriculum domain adaptation for semantic segmentation of urban
  scenes.
\newblock {\em CoRR}, abs/1707.09465, 2017.

\bibitem{zhang14ECCV}
Z.~Zhang, P.~Luo, C.~C. Loy, and X.~Tang.
\newblock Facial landmark detection by deep multi-task learning.
\newblock In {\em ECCV}, pages 94--108, 2014.

\end{thebibliography}
